\documentclass{article}
\usepackage[left=3cm,right=3cm,top=2.5cm,bottom=2.5cm,includeheadfoot]{geometry}
\usepackage{etoolbox}
\usepackage{graphicx}
\usepackage{amsmath,amssymb, amsthm}
\usepackage{dsfont}
\usepackage{enumitem}

\def\cf{\emph{cf.}} 
\newcommand{\ie}{{i.e.}}

\usepackage[utf8]{inputenc}

\usepackage{tabularx}

\usepackage{times}
\usepackage{graphicx}
\usepackage{nicefrac}
\usepackage{MnSymbol}
\usepackage{mathtools}
\usepackage{multirow}
\usepackage{algorithm2e}
\usepackage{bm}
\usepackage{hyperref}
\usepackage{cleveref}
\usepackage{placeins}

\newtheorem{proposition}{Proposition}
\newtheorem{theorem}{Theorem}

\newcommand{\R}{\mathbb{R}}

\newcommand{\eps}{\epsilon}
\newcommand{\opd}{\mathrm{d}}

\newcommand{\sample}{\mathcal S_\epsilon}
\newcommand{\discGrad}[2]{\partial_{(#1,#2)}}
\newcommand{\discHess}[2]{\partial^2_{(#1,#2)}}
\newcommand{\energy}{\mathcal E}
\newcommand{\err}{\mathrm{err}}
\newcommand{\hd}{\mathcal H}
\newcommand{\distortion}{{\gamma}}
\newcommand{\Distortion}{{\Gamma}}

\DeclarePairedDelimiter\abs{\lvert}{\rvert}

\DeclareMathOperator{\av}{av}
\DeclareMathOperator{\Grad}{grad}
\DeclareMathOperator{\grad}{D\!}
\DeclareMathOperator{\Hess}{Hess}
\DeclareMathOperator{\hess}{D^2\!}

\DeclareMathOperator{\tr}{tr}

\begin{document}
\title{Learning low bending and low distortion manifold embeddings\footnote{ © 2021 IEEE. Personal use of this material is permitted. Permission
from IEEE must be obtained for all other uses, in any current or future
media, including reprinting/republishing this material for advertising or
promotional purposes, creating new collective works, for resale or
redistribution to servers or lists, or reuse of any copyrighted
component of this work in other works.}}

\author{
Juliane Braunsmann\thanks{Univerity of M\"unster (\url{j.braunsmann@uni-muenster.de}, \url{benedikt.wirth@uni-muenster.de})} \and Marko Rajkovi\'c \thanks{Institute for Numerical Simulation, University of Bonn (\url{marko.rajkovic@ins.uni-bonn.de}, \url{martin.rumpf@ins.uni-bonn.de})} \and Martin Rumpf\footnotemark[2] \and Benedikt Wirth \footnotemark[1]}
\maketitle

\begin{abstract}
Autoencoders are a widespread tool in machine learning to transform high-dimensional data into a lower-dimensional
representation which still exhibits the essential characteristics of the input.
The encoder provides an embedding from the input data manifold into a latent space which may then be used for further processing.
For instance, learning interpolation on the manifold may be simplified via the new manifold representation in latent space.
The efficiency of such further processing heavily depends on the regularity and structure of the embedding.
In this article, the embedding into latent space is regularized via a loss function that promotes an as isometric and as flat embedding as possible.
The required training data comprises pairs of nearby points on the input manifold together with their local distance and their local 
Fr\'echet average. 
This regularity loss functional even allows to train the encoder on its own. 
The loss functional is computed via a Monte Carlo integration which is shown to be consistent 
with a geometric loss functional defined directly on the embedding map.
Numerical tests are performed using image data that encodes different data manifolds.
The results show that smooth manifold embeddings in latent space are obtained.
These embeddings are regular enough such that interpolation between not too distant points on the manifold
is well approximated by linear interpolation in latent space.
\end{abstract}

\section{Introduction}
\label{sec:intro}
A central task in machine learning is to represent objects in high-dimensional data  manifolds by points in a lower-dimensional hidden latent space. 
Methods in this direction can be split into linear and nonlinear approaches. 
The former include \emph{principal component analysis (PCA)} and \emph{multidimensional scaling (MDS)}~\cite{MaFu11}, examples for the latter are \emph{Isomap}~\cite{TeSiLa00}, \emph{Local Linear Embedding}~\cite{RoSa00} and \emph{Hessian Eigenmaps}~\cite{DoGr03}. 
Such methods take a collection of high-dimensional data points as input and give a collection of low-dimensional vectors as output. 
They rely on neighborhood graphs, and a central part of these methods is usually the computation of a spectral embedding and computation of eigenvalues.

A more recent approach to nonlinear dimensionality reduction are a special type of neural networks called \emph{autoencoders}.
They consist of an \emph{encoder} and a \emph{decoder}. The encoder maps from the high-dimensional ambient space of the data manifold to a low-dimensional Euclidean space, called \emph{latent space}. The decoder maps from latent space back to the ambient space of the data manifold and tries to reproduce the original input data.
In the training phase, the encoder and decoder mapping are determined via the minimization of a loss functional.
The image of a smooth data manifold via a smooth encoder map is a smooth submanifold in the Euclidean latent space.
The assumption that the observed high-dimensional data actually forms a low-dimensional manifold -- the image of the latent manifold under 
the decoder map -- is called the \emph{manifold hypothesis}. 

In \emph{deep manifold learning} one aims at recovering a simpler, low-dimensional latent manifold representation of the data manifold from the observed data via the minimization of 
a loss functional.  In the first place, this loss functional measures the \emph{reconstruction loss} by comparing the input data with its image under the composition of encoder and decoder mapping.
Different strategies have been investigated in addition to the reconstruction loss which favor smoothness of the encoder and decoder mapping and 
thus regularity of the latent manifold.
Among them are methods which promote sparsity~\cite{RaPoChLe07}, contractive autoencoders~\cite{RiViMuGlBe11}, or denoising autoencoders~\cite{ViLaBeMa08}.

Data manifolds often come with a metric 
encoding the cost of local variations on the manifold. 
For sufficiently regular data manifolds it is shown in \cite{ShKuFl18} how to transfer this metric to the latent manifold
and thereby make it \emph{Riemannian}.

Autoencoders also offer the possibility to interpolate between data points by interpolating linearly in the latent space. 
In \cite{BeRaRoGo18} an \emph{adversarial regularizer} was proposed to ensure visually realistic interpolations in latent space.
The adversarial regularizer tries to make the decoding of interpolations in latent space indistinguishable from real data points. Recently, a generalized definition of interpolation via the training of a discriminator was proposed in \cite{OrYaHe20} 
which allows to check that the interpolated point belongs to the original dataset.
While the method in \cite{BeRaRoGo18} relies solely on an adversarial network to discriminate between real data and interpolations, 
the approach in \cite{OrYaHe20} also suggests to include ground truth interpolation data. 
 
A major deficit of autoencoders is that 
they frequently fail to reproduce the statistical input data distribution in the latent space.
In \cite{KaZhSaNa20} \emph{isometric}, \ie, length-preserving encoder maps are used to more accurately push forward distributions from input to latent space.
To this end a loss function based on Shannon-Rate-Distortion theory is proposed.

The loss functional from \cite{AtGrLi20} promotes isometry of the decoder map (by penalizing deviation from a non-orthogonal Jacobian matrix) and that the 
encoder is a pseudo-inverse of the decoder (by enforcing the Jacobians of de- and encoder to be transposes of each other).
In \cite{PeLiDi+20} isometric embeddings in latent space are learned to obtain standardized data coordinates from scientific measurements.
The authors approximate the  Jacobian through normally distributed sampling around each data point (so-called \emph{bursts}), and 
an objective functional measures the lack of orthogonality of the Jacobian via the deviation of the local covariance of bursts from the identity.
In \cite{SchTa19}, with the goal of producing a globally isometric encoder map, a loss functional is proposed which measures the difference between distances in the pushforward metric and distances 
in latent space.  
The training of encoders in  \cite{PaTaBrKi19} is based on
a loss functional which compares Euclidean distances in latent space with geodesic distances on the input manifold.

\paragraph{Our contribution.}
This paper investigates a loss functional for the geometric regularization of the latent manifold. 
Such a loss requires some geometric data of the input manifold in the training.
We follow here a minimalistic approach involving data on distances and averages on the input manifold.
We provide as testbed examples for which this data is explicitly known and easy to compute.
\begin{itemize}
\item We propose to complement a \emph{discrete loss} functional that promotes \emph{isometric embeddings} in latent space
with a discrete \emph{bending loss} functional which prefers as flat as possible embeddings; this combination optionally permits to train the encoder on its own.\\[-4ex]
\item To train a corresponding autoencoder network we consider training data consisting of \emph{triplets} of two input data points and their 
\emph{Fr\'echet average} together with the distance between the two data points.
\\[-4ex]
\item Unlike other isometry promoting approaches we do not approximate the \emph{Jacobian of the encoder}
(via backpropagation or complete correlation of all nearest neighbors) but use simple \emph{Monte\,Carlo point sampling}.
\\[-4ex]
\item Matching the theory of isometric maps, our numerical experiments confirm that the bending loss significantly increases \emph{smoothness} of the resulting latent space manifold
over a pure isometry loss.  Also we find that the decoder maps 
\emph{linear interpolation in latent space} to reasonable interpolations on the data manifold.
\\[-4ex]
\item We demonstrate that our discrete loss functional is consistent with a 
\emph{well-posed, continuous limit functional} on encoder maps from a smooth 
Riemannian data manifold into latent space. 
 
\end{itemize} 

The paper is structured as follows.
\Cref{sec:discreteloss} derives the new regularization loss and discusses its Monte Carlo limit for dense sampling of 
point pairs.
The limit functional of this Monte Carlo limit for vanishing distance between the point pairs is introduced in \cref{sec:limit}, and we prove existence of minimizers. 
As a proof of concept, we train autoencoders on three image datasets representing a priori known data manifolds.
\Cref{sec:experiment} describes the autoencoder set up, the experimental datasets, and the numerical results.

\section{A low bending and low distortion regularization for encoders}
\label{sec:discreteloss}
Consider a smooth compact $m$-dimensional Riemannian manifold $(M,g)$ possibly with boundary, of which we have samples available.
To avoid technical details the theory will be presented only for input manifolds without boundary.
We assume that $M$ is embedded in some very high-dimensional space $\R^n$, for instance the space of images 
with $n$ pixels.
The aim is now to compute an embedding of  $M$ into Euclidean space $\R^l$ 
(called the \emph{latent space}),
where we think of the dimension $l$ as being only moderately larger than $m$
(for instance $l=2m$ so that the existence of a smooth embedding is guaranteed by Whitney's embedding theorem).
Such a representation is often learned from samples by training an autoencoder, which is 
a pair of maps 
\begin{align} 
\phi:M\to\R^l,\text{ }  \psi:\R^l\to\R^n \text{ with }  \psi(\phi(x))\approx x \text{ for all } x\in M.
\end{align}
The image $\phi(M)$ of $M$ is called \emph{latent manifold}.
The autoencoder functions $\phi$ and $\psi$ are implemented as deep neural networks.
An appropriate structure and regularity of the embedding $\phi$ into latent space 
is known to aid downstream tasks such as classification, Riemannian interpolation and extrapolation, clustering or anomaly detection.
For this reason we aim for a natural, geometrically inspired regularizing loss function for the encoder $\phi$.

From the viewpoint of downstream processing,
the nicest embedding $\phi: M \to \R^l$ would of course be an isometric embedding into an affine subspace of $\R^l$.
This would identify $M$ as isometric to flat Euclidean space
so that any downstream processing on $M$ can be performed with the simplicity and efficiency immanent to Euclidean space.
Of course, such an embedding is usually prevented by the intrinsic or the global geometry of $M$,
nevertheless one may try to get as close as possible to an {\bf I}sometric and {\bf F}lat embedding, at least locally.
Hence we suggest the following two simple objectives for any two not too distant points $x,y\in M$:\\[-1\baselineskip]
\begin{itemize}
\item[({\bf I})]
The intrinsic Riemannian distance between $x$ and $y$ in $M$ 
should differ as little as possible from the Euclidean distance between the 
latent codes $\phi(x)$\,and\,$\phi(y)$.\\[-1\baselineskip]
\item[({\bf F})]
A (weighted) average between $x$ and $y$ in $M$ 
should deviate as little as possible from the (weighted) Euclidean average between $\phi(x)$ and $\phi(y)$.\\[-1\baselineskip]
\end{itemize}
If the first objective is only applied to infinitesimally close points $x,y\in M$, it is nothing else than asking for an \emph{isometric embedding},
as extensively pursued in the literature (\cf \cref{sec:intro}).
It ensures that $\phi$ embeds $M$ in latent space with low distortion.
However, asking for isometry alone is highly questionable from the mathematical point of view
since the family of isometric embeddings is very large and contains quite irregular elements
(Nash--Kuiper embeddings are in general only H\"older differentiable).
Therefore, in (\textbf{I}) and (\textbf{F}) we go beyond this infinitesimal isometry viewpoint:

$\bullet$ The isometry  objective (\textbf{I}) asks that the \emph{intrinsic} distances between $x$ and $y$ in $M$ are approximated by the \emph{extrinsic} distances between $\phi(x),\phi(y)\in\R^l$ in latent space (rather than the intrinsic distances in the latent manifold $\phi(M)$, which would define an isometric embedding).

$\bullet$ The flatness/bending objective (\textbf{F}) enforces some second order low bending regularity or flatness on $\phi$ by requiring that the \emph{geodesic interpolation} between $x$ and $y$ in $M$ is well approximated by extrinsic \emph{linear interpolation} in the latent space $\R^l$.

\paragraph{A low bending and low distortion loss.}
Denote the geodesic distance between two points $x,y\in M$ by $d_M(x,y)$ and their geodesic average by $\av_M(x,y)$.
As input data to the training or optimization of the encoder $\phi$ we consider a sample $\sample\subset\{(x,y)\in M\times M\,|\,d_M(x,y)\leq\epsilon\}$
of pairs $(x,y)\in M\times M$ of nearby points
together with $d_M(x,y)$ and $\av_M(x,y)$.
For a sufficiently small fixed locality radius $\epsilon>0$, the unique existence of the geodesic average is ensured.
Our proposed loss function to regularize the encoder $\phi$ then reads
\begin{equation}
E^{\sample}(\phi) =
\frac1{|\sample|}\sum_{(x,y)\in\sample}\left(\distortion(\discGrad{x}{y}\phi) + \lambda \; |\discHess{x}{y}\phi|^2\right)
\end{equation}
with first and second order difference quotients 
\begin{equation}
\discGrad{x}{y}\phi=\frac{\phi(y)-\phi(x)}{d_M(x,y)}, ~\discHess{x}{y}\phi=8\frac{\av_{\R^l}(\phi(x), \phi(y))-\phi(\av_M(x,y))}{d_M(x,y)^2},
\end{equation}
where $\distortion(s) = |s|^2+|s|^{-2}-2$, $\av_{\R^l}(a,b)=(a+b)/2$ denotes the linear average in $\R^l$, and $\lambda>0$.
Note that the first term in $E^{\sample}$ has
a strict minimum for  $|\discGrad{x}{y}\phi|=1$.
This term thus promotes $\abs{\phi(x)-\phi(y)}=d_M(x,y)$ and thus low distortion and approximate isometry.
The second term in $E^{\sample}$ penalizes the deviation of intrinsic averages on $\phi(M)$ from extrinsic ones in $\R^l$.
Note that this does not only penalize bending or any extrinsic curvature of $\phi(M)$ in $\R^l$,
but in addition it also penalizes deviation of the inplane parameterization of $\phi(M)$ from a linear one (\cf the corresponding 
remark for $\Hess\phi$ in the next section).
Examples of how to compute $\av_M$ and $d_M$ for image input data include the corresponding methods 
from the theories of LDDMM \cite{Yo10}, metamorphosis \cite{TrYo05a}, or optimal transport \cite{CuPe19}.
In our testbed we purposely used low-dimensional manifolds where $\av_M$ and $d_M$ are explicitly known.

\paragraph{The Monte Carlo limit for dense sampling.}
Assuming that $\sample$ is drawn uniformly from $M\times M$ (subject to the locality condition),
our loss function $E^{\sample}$ is up to $\mathcal{O}(\epsilon)$ the Monte Carlo integration
of the energy
\begin{align}\label{eqn:loss}
   \energy^\eps(\phi)\!=\!\strokedint\limits_M\!\! \strokedint\limits_{B_\eps^M\!(x)} \hspace{-1.7ex} \distortion(\discGrad{x}{y}\phi)\! +\! \lambda \, |\discHess{x}{y}\phi|^2 \opd V_g(\!y\!)\opd V_g(\!x\!),
\end{align}
where $B_\eps^M(x)$ denotes the geodesic $\eps$-ball in $M$, centered at $x$,
and where $\strokedint\ldots\opd V_g$ denotes the mean with respect to the Riemann--Lebesgue volume measure on $M$ (the index $g$ indicates the Riemannian metric).
As for the discrete functional $E^{\sample}$, the energy $\energy^\eps$ penalizes deviation from isometry and from intrinsically and extrinsically flat embeddings.

The energy $\energy^\eps$ is rigid motion invariant by construction, \ie, composition of $\phi$ with a rigid motion does not change the energy. 
However, even apart from this invariance one cannot expect uniqueness of minimizers due to the nonconvexity of the first integrand.
Whenever $M$ is intrinsically flat and homeomorphic to the $m$-disc (or at least globally compatible with an embedding into an $m$-dimensional Euclidean space), there is a unique minimizer of $\energy^\eps$, though, as stated in the following proposition.
\begin{proposition}[unique embedding of intrinsically flat discs]
If $M$ is the flat $m$-disc $D^m$, the unique minimizer (up to rigid motion) of $\energy^\eps$ is $\phi:M\ni x\mapsto(x,0,\ldots,0)\in\R^l$.
\end{proposition}
\begin{proof}
It is straightforward to check $\energy^\eps\geq0$ as well as $\energy^\eps(\phi)=0$ so that $\phi$ is a global minimizer.
The uniqueness up to rigid motion then follows from the fact
that fixing $\phi$ at $m+1$ points $x_0,\ldots,x_m$ uniquely determines $\phi$ at all points $x$ within the convex hull of $x_0,\ldots,x_m$
since such $x$ can be represented as (limit of) iterated averages of $x_0,\ldots,x_m$
so that $\phi(x)$ must be the (limit of the) corresponding iterated averages of $\phi(x_0),\ldots,\phi(x_m)$
(that one may take limits of $\phi$ follows from the condition $\discGrad{x}{y}\phi=1$ for all close enough $x,y$).
However, fixing $m+1$ points with prescribed distances just fixes a rigid motion.
\end{proof}

Note that this property of recovering flat embeddings may be quite relevant in applications
as generative image manifolds were noticed in \cite{ShKuFl18} to have almost no curvature.

\paragraph{Affinely invariant loss functions.}
In several applications one may already be content with a nice embedding $\phi$ that is specified only up to an affine transformation (rather than a rigid motion).
Indeed, one may want to abandon isometry in favor of improving the approximation of geodesic averages by linear averages.
This raises the question whether one can replace the integrand in \eqref{eqn:loss} by some function $f(\discGrad{x}{y}\phi,\discHess{x}{y}\phi)$
which is invariant under left composition of $\phi$ with invertible affine (and not just rigid) transformations.
Yet, if $\discGrad{x}{y}\phi$ and $\discHess{x}{y}\phi$ are non-parallel
there is always an affine transform that maps $\discGrad{x}{y}\phi$ onto $(1,0,\cdots)$ and $\discHess{x}{y}\phi$ onto $(0,1,0,\cdots)$
so that necessarily $f$ is of the form $f(a,b)\equiv c$ for $a$, $b$ non-parallel and $f(a,b) = h(s)$ for $b=sa$
with some constant $c$ and function $h:\R\to\R$.
Thus, the regularizing properties would be lost.
An alternative could be to just penalize $|\discHess{x}{y}\phi|/|\discGrad{x}{y}\phi|$.
Though not affinely invariant, it still encodes that geodesics should be close to linear interpolation without any competing isometry constraints
($|\discGrad{x}{y}\phi|$ in the denominator is needed for scale invariance and prevents a collapse to 
$\phi(M)=0$).
Our loss $\energy^\eps$ controls this flatness measure due to
\begin{equation}
2\sqrt\lambda|\discHess{x}{y}\phi|/|\discGrad{x}{y}\phi| \leq 
|\discGrad{x}{y}\phi|^{-2} + \lambda\, |\discHess{x}{y}\phi|^2\,.
\end{equation}
Replacing $\energy^\eps(\phi)$ with $\strokedint_M\strokedint_{B_\epsilon^M(x)}|\discHess{x}{y}\phi|/|\discGrad{x}{y}\phi|\opd y\, \opd x$ would be infeasible, though:
a straightforward calculation shows that a minimizing sequence of embeddings of the cylinder $M=S^1\times[0,1]$ into $\R^l$ would be cylinders of vanishing radius and diverging length.

\section{The limit of vanishing locality radius}
\label{sec:limit}
An appropriate structure and regularity of an embedding $\phi:M\to\R^l$ can also be promoted by a purely local functional.
Below we present a natural candidate and identify it as a consistent limit of $\energy^\eps$ when $\eps\to 0$ under smoothness assumptions on the embedding.

\paragraph{A purely local low bending and low distortion loss.}
The \emph{Riemannian gradient (Jacobian)} $\Grad\phi(x)\in(T_xM)^l$ of $\phi$ is defined (denoting standard differentiation of smooth extensions onto $\R^n$ by $\grad$\,) via the identity
\begin{equation}
\grad\phi_j(x)(v)=\tfrac{\mathrm{d}}{\mathrm{d}t}(\phi_j\circ\exp_x)(tv)\vert_{t=0}=g(\Grad\phi_j(x),v)
\end{equation}
for all $v\in T_xM$, where $\exp_x:T_xM\to M$ denotes the Riemannian exponential map in $x$.
An isometric embedding $\phi:M\to\R^l$ is characterized
by $\Grad\phi(x)$ being orthogonal in any point $x\in M$.
Thus, deviation from an isometric embedding manifests as non-\emph{unit singular values of $\Grad\phi(x)$}.
Similarly, extrinsic bending of the embedding $\phi(M)$ manifests as a non-\emph{vanishing Riemannian Hessian} $\Hess\phi$ of $\phi$,
where the Riemannian Hessian at $x$ is the linear operator 
\begin{equation}
\Hess\phi(x):T_xM\to(T_xM)^l, \quad \Hess\phi(x)(v)=(\nabla_v\Grad\phi_j)_{j=1,\ldots,l}
\end{equation}
for $\nabla$ the Levi-Civita connection and $\nabla_v$ the covariant derivative in direction $v$ \cite[Chp.\,5]{Absil09}.
The associated quadratic form $\hess\phi(x):T_xM\times T_xM\to\R^l$ is
\begin{align}
\hess\!\phi_j(x)(v,v)
=\tfrac{\mathrm{d}^2}{\mathrm{d}^2t}(\phi_j\circ\exp_x(tv))\vert_{t=0}
= g(\Hess\phi_j(x)(v),v),
\end{align}
where the Riemannian metric $g$ on $(T_xM)^l\times T_xM$ is applied componentwise, \ie, we use the notation
$g(A,v) = (g(A_j,v))_{j=1,\ldots, l}$ for a matrix whose rows $A_j$ are tangent vectors.
A natural loss function to promote low distortion and low bending embeddings thus reads
\begin{align}
\energy(\phi)
\!=\!\strokedint_M \Distortion(\Grad\phi(x))
+\tfrac\lambda{2} \|\Hess\phi(x)\|_F^2\opd V_g(x), \label{eq:limitE}
\end{align}
where $\|A\|_F^2=\tr(A^*A)$ is the Frobenius norm of an operator $A$
and $B \mapsto \Distortion(B)$ is an \emph{admissible} function, \ie, a nonnegative function depending only on the singular values of $B$ and being zero if and only if all of them are one.

Let us remark that the orthogonal projection of $\Hess\phi(x)$ onto the normal bundle $[T_{\phi(x)}\phi(M)]^\perp$ of
$\phi(M)$ is the second fundamental form of $\phi(M)$ pulled back onto $T_xM$.
If $m=2$ and $l=3$, this is also known as the \emph{Weingarten map} or \emph{shape operator relative to $T_xM$}.
In addition to penalizing this second fundamental form, which indicates extrinsic bending of $\phi(M)$,
our functional $\energy$ also penalizes the tangential components of $\Hess\phi$.

\paragraph{Reformulation with directional derivatives.}
The above energy can be rewritten in terms of averages.
To this end, let $S^{m-1}$ denote the unit sphere in $T_xM$ (the base point $x$ will be clear from the context),
and let $\hd^{m-1}$ denote the $(m-1)$-dimensional Hausdorff measure in $T_xM$.
\begin{proposition}[double integral representation of $\energy$]
The choice $\Distortion:(T_xM)^l\to[0,\infty]$,
\begin{equation}
\textstyle
\Distortion(B)=\strokedint_{S^{m-1}}\distortion(g(B,v))\; \opd\hd^{m-1}(v)
\end{equation}
is admissible in the above sense and leads to the representation
\begin{align}
\energy(\phi)
=\strokedint_M\strokedint_{S^{m-1}} & \distortion(\grad\phi(x)(v)) 
+ \lambda |\hess\phi(x)(v,v)|^2 \opd\hd^{m-1}(v)\opd V_g(x).
\end{align}
\end{proposition}
\begin{proof}
Let $B$ have singular values $\sigma_1,\ldots,\sigma_m$ and left and right singular vectors $w_1,\ldots,w_m\in\R^l$, $v_1,\ldots,v_m\in T_xM$,
then 
\begin{equation}
|g(B,v)|^2=|\sum_{i=1}^m\sigma_ig(v_i,v)w_i|^2 =\sum_{i=1}^m\sigma_i^2g(v_i,v)^2.	
\end{equation}
Inserting this expression into $\Distortion$, 
one sees that the integral makes the expression independent of the orthonormal frame $v_1,\ldots,v_m$
so that $\Distortion(B)$ indeed only depends on the singular values of $B$.
Furthermore, $\Distortion$ is nonnegative since its integrand is, 
and it is zero if and only if $|g(B,v)|=1$ for all $v\in S^{m-1}$ in $T_xM$,
which by the above is equivalent to all singular values being one.
Analogously one shows that
\begin{align}
\strokedint_{S^{m-1}}\sum_{j=1}^l|\hess\phi_j(x)(v,v)|^2\opd\hd^{m-1}(v)
=\sum_{j=1}^l \sum_{i=1}^m(\sigma^j_{i})^2\!\!\strokedint_{S^{m-1}}g(v,e)^2\opd\hd^{m-1}(v)
=\tfrac12\|\Hess\phi(x)\|_F^2
\end{align}
for the eigenvalues $\sigma^j_{i}$ of $\Hess\phi_j(x)$ and some arbitrary $e\in S^{m-1}$ using that the integral inside the sum is $\tfrac12$.
\end{proof}
\paragraph{Identification as limit for vanishing locality radius.}
It turns out that for $\eps\to0$ our loss $\energy^\eps$ approximates $\energy$,
which thus gives a simple interpretation of $\energy^\eps$ in terms of first and second order derivatives of $\phi$.

\begin{theorem}[limit energy for vanishing $\epsilon$]\label{thm:concistency}
$\energy^\eps$ is a consistent approximation of $\energy$
in the sense $\energy^\eps(\phi) = \energy(\phi) + \mathcal O(\eps\|\phi\|_{C^3})$.
\end{theorem}
\begin{proof}
   For $\eps$ small enough, the Riemannian exponential $\exp_x$ defines a diffeomorphism between the $\eps$-ball $B_\eps(0) \subset T_xM \cong \R^m$ and $B_\eps^M(x)$
   with inverse denoted by $\log_x$.
   For any measurable function $f_x: M \to \R$ we then have
   \begin{equation}\label{exp-trafo}
      \strokedint_{B_\eps^M(x)} f_x(y)\opd V_g(y) = \strokedint_{B_\eps(0)} f_x(\exp_xw)\opd (\log_x^*V_g)(w),
   \end{equation}
where $\log_x^*V_g$ is the pushforward measure of $V_g$ under $\log_x$. The Lebesgue density of $\log_x^*V_g$ at $w\in B_\eps(0)$ is known to have the expansion $1+\mathcal O(|w|^2)$
   (the constant depends on the Ricci curvature, \cf \cite{Agrachev19}).
This can be used together with the transformation formula to get
   \begin{align}
\strokedint\limits_{B_\eps(0)}\hspace{-0.5em}f_x(\exp_xw)\opd (\log_x^*V_g)(w) &=  
\strokedint\limits_{B_1(0)}\hspace{-0.5em}f_x(\exp_x(\eps w))(1 + \mathcal O(\eps^2))\,\opd w \\ \nonumber
&=\strokedint\limits_{S^{m-1}}\hspace{-0.5em}\int\limits_0^1\hspace{-1ex} f_x(\exp_x(\eps rv))(1\! +\! \mathcal O(\eps^2))mr^{m-1}\opd r\opd\hd^{m-1\!}(v).
   \end{align}
Now we consider the cases $f_x(y) = \distortion(\discGrad{x}{y}\phi)$ as well as $f_x(y)=|\discHess{x}{y}\phi|^2$.
Letting $y=\exp_x(\eps rv)$ and abbreviating $\theta(t)=\phi(\exp_x(tv))$, Taylor expansion yields
\begin{align}
\discGrad{x}{y}\phi
&=\frac{\theta(\eps r)-\theta(0)}{r\eps}
=\theta'(0) + \mathcal O(r\eps)
\quad\text{ and}\\
\discHess{x}{y}\phi
&=8\frac{\tfrac{1}{2}(\theta(0)+\theta(\eps r))-\theta(\tfrac{\eps r}{2})}{r^2\eps^2}
=\theta''(0) + \mathcal O(r\eps).
\end{align}
Now by the definition of gradient and Hessian we have $\theta'(0)=\grad\phi(x)(v)$ and 
$\theta''(0)=\hess\phi(x)(v,v)$.
The proof is concluded by inserting these estimates in $f_x$ and noting that the constants of all error terms in $\eps$ depend on the manifold $M$ and on (at most) third derivatives of $\phi$.
\end{proof}
\paragraph{Existence of optimal geometric embeddings.}
Let us now establish the existence of minimizers to $\energy$.
First, we observe that the energy $\energy$ is  well-defined on all of $H^2(M)$, where the Sobolev space 
$H^2(M)$ is defined as the closure of all smooth functions under the norm $\|\phi\|_{H^2(M)}$ with 
\begin{equation}\label{eq:sobolev_norm}
\|\phi\|_{H^2(M)}^2
=\sum_{j=1}^m\int_M|\phi_j|^2
+g(\Grad\phi_j,\Grad\phi_j)
+g(\Hess\phi_j,\Hess\phi_j)\; \opd V_g .
\end{equation} 
For further details on Sobolev spaces on (compact) manifolds we refer to \cite{Hebey96}.
Due to the rigid motion invariance of $\energy$ we may without loss of generality restrict $\energy$ to the subspace $\dot H^2(M)$ of $H^2$-functions with zero average.
\begin{theorem}[existence of a minimizer]
Let $M$ be smooth, compact.
If there exists $\phi\!\in\!\dot H^2(M)$  with $\energy(\phi)<\infty$,
then $\energy$  has a minimizer in $\dot H^2(M)$.
\end{theorem}
If $l\geq 2m$ the condition is always fulfilled since by Whitney's embedding theorem there exists a smooth embedding which, due to the compactness of $M$, may be chosen such that it has finite energy.
\begin{proof}
We apply the direct method in the calculus of variations.
By our assumption there exists  a minimizing sequence $(\phi^k)_{k=1,2,\ldots} \subset \dot H^2(M)$, which we 
suppose to converge monotonically to $\inf\energy<\infty$.
Since $\Distortion(\Grad\phi)\geq C|g(\Grad\phi,\Grad\phi)|-2$ and 
$\|\Hess\phi\|_F^2\geq C|g(\Hess\phi,\Hess\phi)|$ for some constant $C>0$,
the second and third summand of 
\eqref{eq:sobolev_norm} are uniformly bounded for all $\phi^k$.
By Poincar\'e's inequality this implies uniform boundedness of $\phi^k$ in $\dot H^2(M)$.
By reflexivity of $\dot H^2(M)$, there exists a weakly convergent subsequence (still indexed by $k$) with 
limit $\phi$ in $\dot H^2(M)$.
Convexity of the map $A\to \|A\|_F^2$ then implies 
$\liminf_{k\to\infty}\strokedint_M\|\Hess\phi^k\|_F^2\opd V_g\geq\strokedint_M\|\Hess\phi\|_F^2\opd V_g$.
Furthermore, by Rellich embedding, $\Grad \phi^k$ already converges strongly to $\Grad \phi$ in $L^2(M)$  
and up to selection of another subsequence even pointwise almost everywhere.
Fatou's lemma then implies 
$\strokedint_M \Distortion(\Grad \phi)\opd V_g\leq\liminf_{k\to\infty}\strokedint_M \Distortion(\Grad \phi^k) \opd V_g$. 
Thus, we obtain lower semi-continuity of the energy, \ie,
 $\energy(\phi)\leq\liminf_{k \to \infty} \energy(\phi^k)=\inf\energy$, which establishes the claim.
\end{proof}
Just as for $\energy^\eps$, the minimizer (modulo a rigid motion) is in general not unique due to the isometry promoting term.

\section{Numerical experiments}
\label{sec:experiment}
In what follows, similarly to \cite{DoGr05}, we consider image data that implicitly represent three different manifolds:\\[-4ex]
\begin{itemize}
\item[(G)] images of anisotropic Gaussians which are rotated, scaled and translated, representing a cylinder $S^1\times [a,b]\times[c,d]^2$,\\[-4ex]
\item[(S)] shadows of a sundial with the sun or light source shining from all possible directions, representing
the upper hemisphere $S^2 \cap \{x_3\geq0\}$ (\cf \cite{OrYaHe20}), \\[-4ex]
\item[(R)] orthogonal projections of a rotated 3D object, representing $SO(3)$.  \\[-4ex] 
\end{itemize}
\paragraph{Datasets.}
We consider an image resolution of $64\times64$.  
The images are generated as follows.

{\it (G) Anisotropic Gaussians.}
We consider rotations, scalings, and translations of a cut off Gaussian of fixed aspect ratio, with
parameters  $(\alpha,s,x) \in M=S^1\times [a,b]\times[c,d]^2$ with distance $$d((\alpha, s, x_1, x_2), (\alpha', s',  x_1',  x_2'))^2 = d_{S^1}(\alpha,\alpha')^2 + \abs{s-s'}^2 + \abs{x_1-x'_1}^2 + \abs{x_2-x'_2}^2,$$ where $d_{S^1}$ is the geodesic distance on $S^1$.
The data is similar to the DSprites dataset in \cite{dsprites17}.

\begin{figure}\centering
\includegraphics[width=0.7\textwidth]{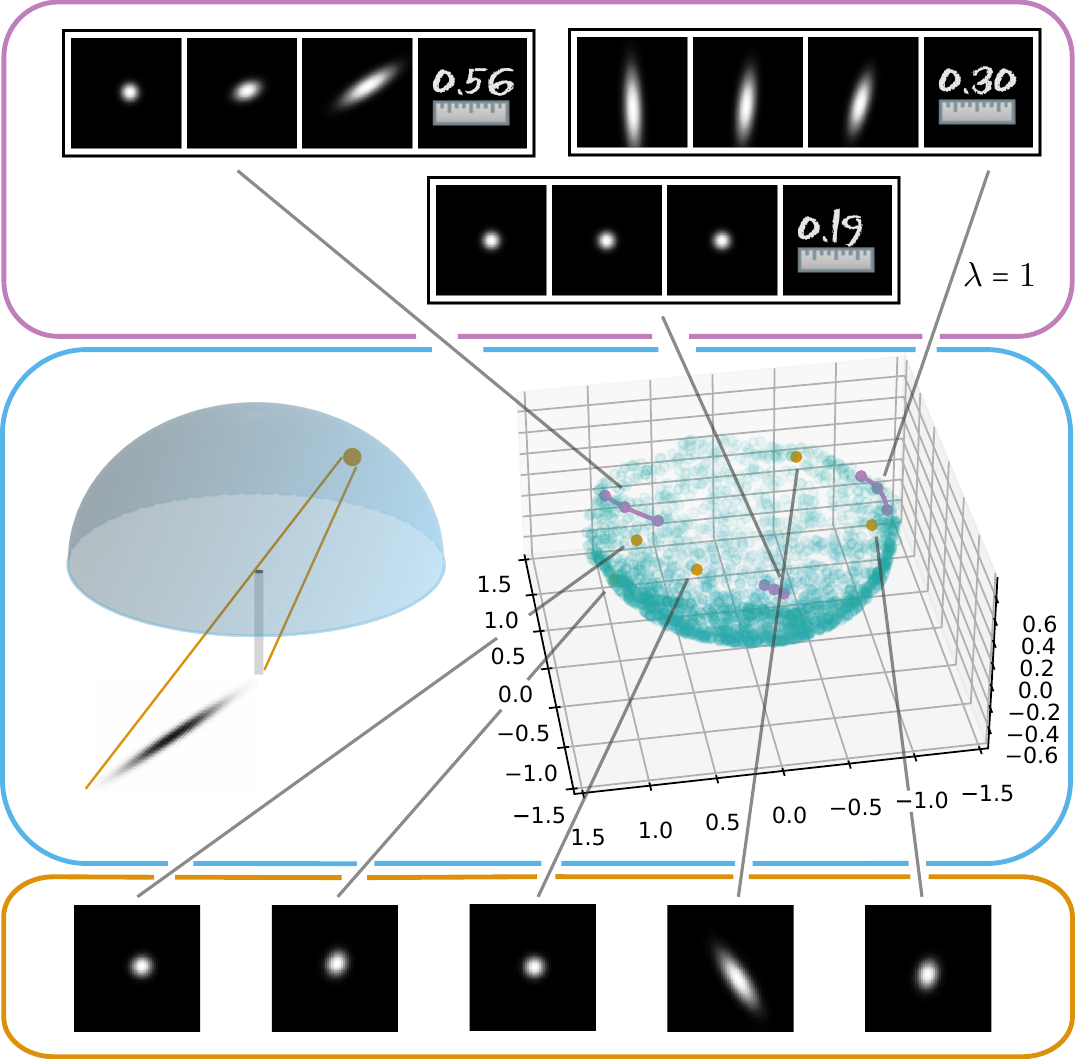}
\caption{Results for the sundial dataset (S).
   The top box shows selected training data (image pairs, geodesic average, distance).
   The middle box shows a sketch of the sundial configuration and the latent manifold $\phi(M)$ projected into $\R^3$ via PCA. 
   The bottom box shows decoder outputs for the orange points in latent space.}
   \label{fig:masterpiece}
\end{figure}

{\it (S) Sundials.}
Inspired by \cite{OrYaHe20}, we generate images parametrized by the upper hemisphere $M=S^2\cap \{x_3 \geq 0\}$ by casting a shadow of a vertical rod on a plane. 
Contrary to \cite{OrYaHe20} we do not render these images with a 3D engine, but instead simply approximate the shadows by Gaussians (\cf \cref{fig:masterpiece}): 
a point $x\in M$ is first mapped onto the plane by drawing the line through $x$ and the rod tip, intersecting the plane at some $y \in \R^2$. 
We then use a Gaussian function with variance $\abs y$ in direction $y$ and a fixed small variance in the orthogonal direction, centered at $y/2$.  
As distance on $M$ we use the geodesic distance on $S^2$, $d(x,x')=\arccos(x^Tx')$.

{\it (R) Rotated 3D objects.}
We generate images by rotating a camera pointing at a three-dimensional object,
Spot the cow\footnote{https://www.cs.cmu.edu/~kmcrane/Projects/ModelRepository/}.
We use \texttt{Pytorch3D} \cite{ravi2020pytorch3d} to render the images during training.
As distance on $M=SO(3)$ we use the geodesic distance computed via quaternions as $d(q_1, q_2)=\arccos(\abs{ q_1 \cdot q_2})$ \cite{Hu09}.

\paragraph{Autoencoder architecture.}
The used architecture is as in \cite{BeRaRoGo18}, however, we used larger input images and a smaller latent dimensionality. The encoder consists of a first layer of $1\times 1$ convolutions with $16$ output channels, followed by blocks consisting of two consecutive $3\times 3$ convolutions with unit stride, zero padded such that the input and output width are equal, and $2\times 2$ average pooling. Each of the convolutional layers in the block is followed by a leaky ReLU nonlinearity with  slope $-0.01$. The first convolution in each block doubles the output channels. 
The final convolution is not followed by a nonlinearity and has only 1 (4 for (R)) output channel(s). The number of convolutional blocks determines the size of the latent code: we used $4$ blocks with input images of size $64\times 64$ ($64\times 64\times 3$ for (R)), resulting in a latent code of size $16$ ($64$ for (R)). The decoder also consists of consecutive blocks of two $3\times 3$ convolutions with leaky ReLU nonlinearities, where each block is followed by $2\times 2$ nearest neighbor upsampling. The final convolutional layer is again not followed by a nonlinearity and has 1 output channel. We use Kaiming initialization \cite{HeZhReSu15}, \ie, all convolutional weights are initialized as zero-mean Gaussian random variables with standard deviation $\sqrt{2} / \sqrt{\texttt{fan\char`_in}(1+0.01^2)}$ for \texttt{fan\char`_in} the layer input dimension, and all biases are initialized with zeros. 
For training, we use the Adam optimizer \cite{KiBa15} with learning rate $0.0001$ and default values for $\beta_1, \beta_2$ and $\varepsilon$. The training data are triplets of images plus a distance value (\cf \cref{fig:masterpiece} top),  $$(x,y,\av_M(x,y),d_M(x,y))\,,$$ with $x,y\in\sample\subset M$, $\av_M(x,y)$ the geodesic average of $x,y$ in $M$ and $d_M(x,y)$ their geodesic distance.
This input allows to compute the ingredients  $\discGrad{x}{y}\phi$  and $\discHess{x}{y}\phi$ of the loss functional $E^{\sample}(\phi)$ defined in 
\cref{sec:discreteloss}. 

\paragraph{Smooth embedding and reliable reconstruction.}
\Cref{fig:masterpiece} summarizes our approach and its result at one glance for dataset (S);
the top box shows examples of training triplets plus distances, 
the middle box visualizes the obtained manifold embedding $\phi(M)$,
and the bottom box displays reconstructions of input images by the full autoencoder.
\begin{figure}[h]
\begin{center}
   \setlength{\tabcolsep}{10pt}
   \begin{tabularx}{\textwidth}{rrcc}
      $\lambda=0$ & $\lambda=5$ \\[-0.5cm]
      \includegraphics[width=0.44\textwidth]{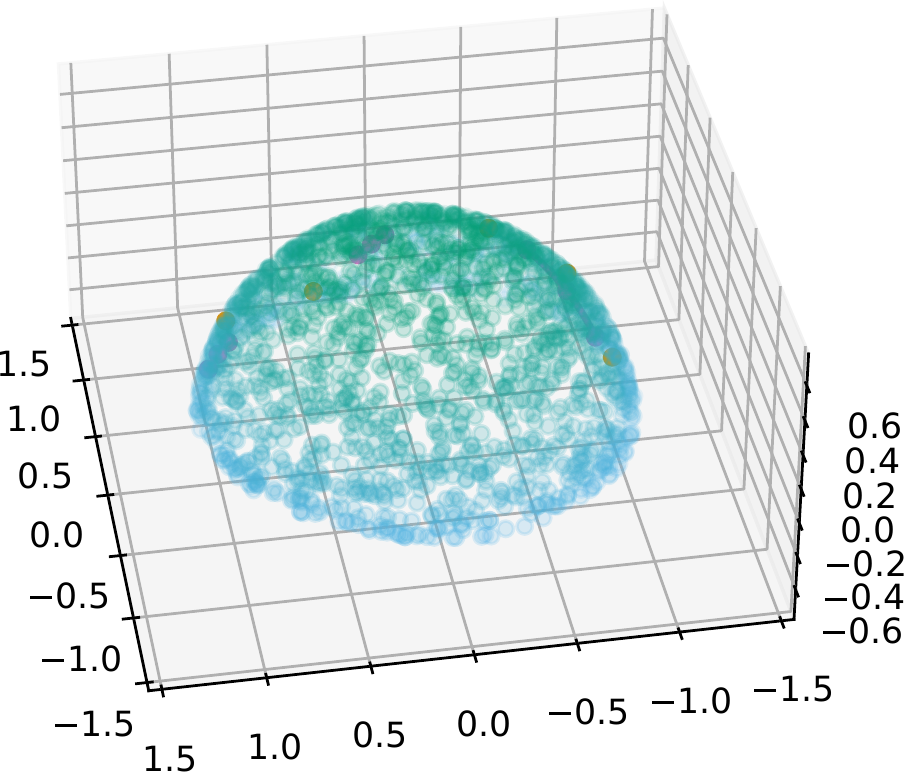} &
      \includegraphics[width=0.44\textwidth]{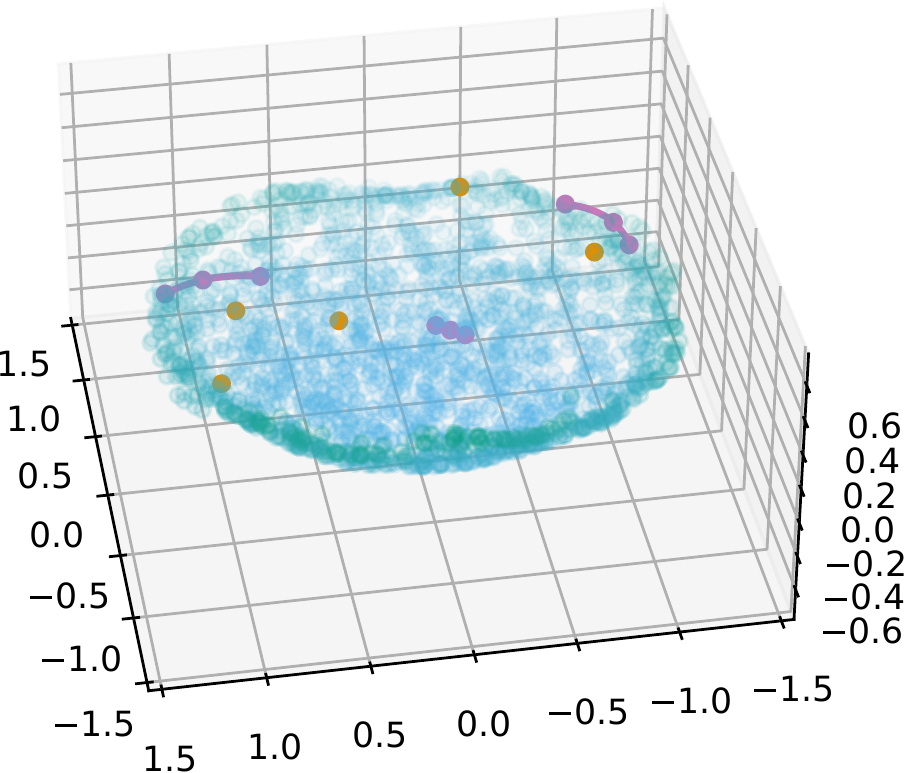} &
   \end{tabularx}
\end{center}
\caption{Latent manifold $\phi(M)$ for sundial dataset (S), obtained for different values of flatness weight $\lambda$
(colored points as in \cref{fig:masterpiece}). 
Due to rigid motion and symmetry invariance of $E^{\sample}$, the latent manifold appears in orientations different from \cref{fig:masterpiece}.}
\label{fig:sundials-lambda}
\end{figure}

\begin{figure}[ht]
\centering
\includegraphics[]{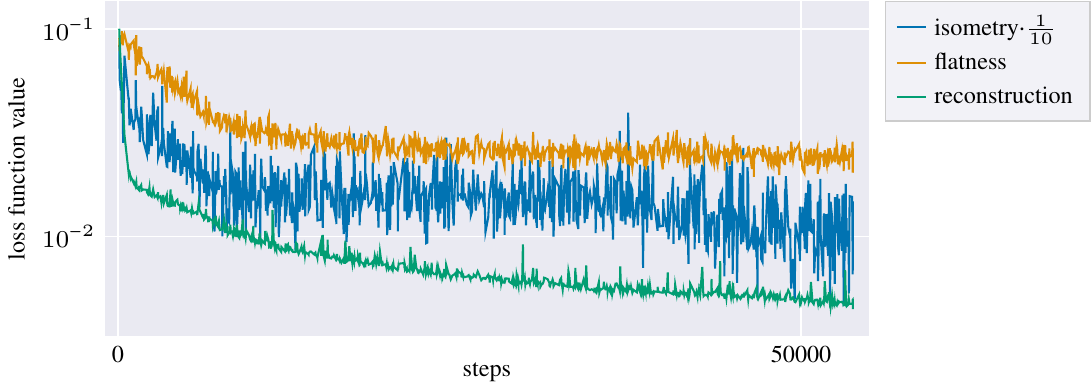}   
\caption{Temporal evolution of the three loss components for dataset (R) of rotated objects (logarithmic $y$-axis, value of isometry loss scaled down by factor 10).
Per optimization step a batch of 128 images is processed.}
\label{fig:training}
\end{figure}
\begin{figure}
\centering
\includegraphics[]{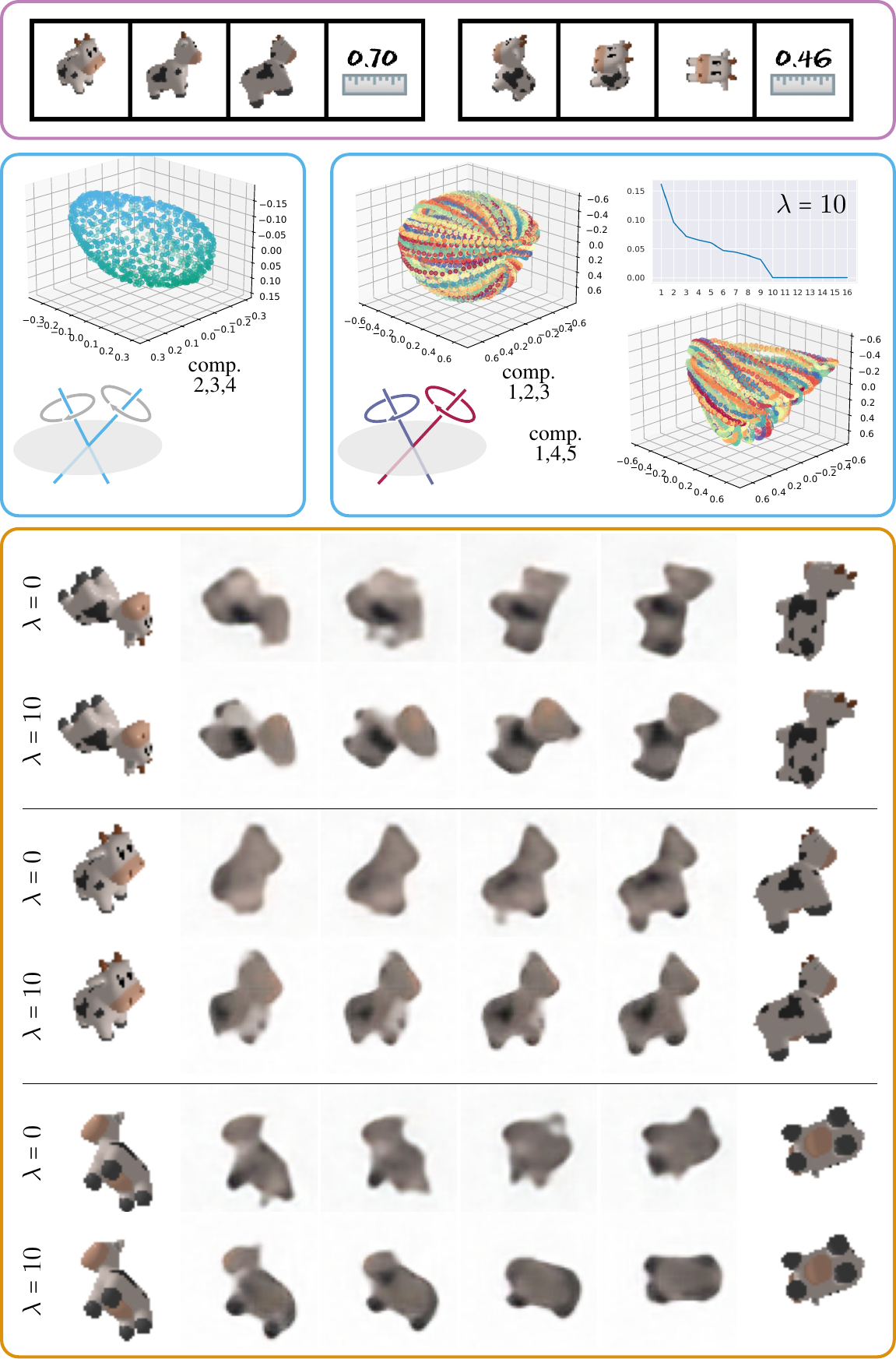}
\caption{Results of our method for the rotated objects dataset (R).
   The middle boxes show projections of the obtained latent manifold $\phi(M)$:
   a submanifold for fixed rotation angle around all possible axes in $S^2$ (left)
   and all of $\phi(M)$ with rotations around the same axis in same color (right, once taking principal components 1, 2, 3, once 1, 4, 5).
   The additional curve illustrates the standard deviation along the principal components in latent space, indicating that 9 Euclidean dimensions are used for the embedding.
   The bottom box shows decoder outputs for linear interpolation in latent space between the codes of the first and last image.
   Such interpolation becomes feasible for higher bending weight $\lambda$, \emph{even though the decoder was not trained for such codes.}}
   \label{fig:cows-qualitative}
\end{figure}
\begin{figure}
\centering
\includegraphics[width=0.6\textwidth]{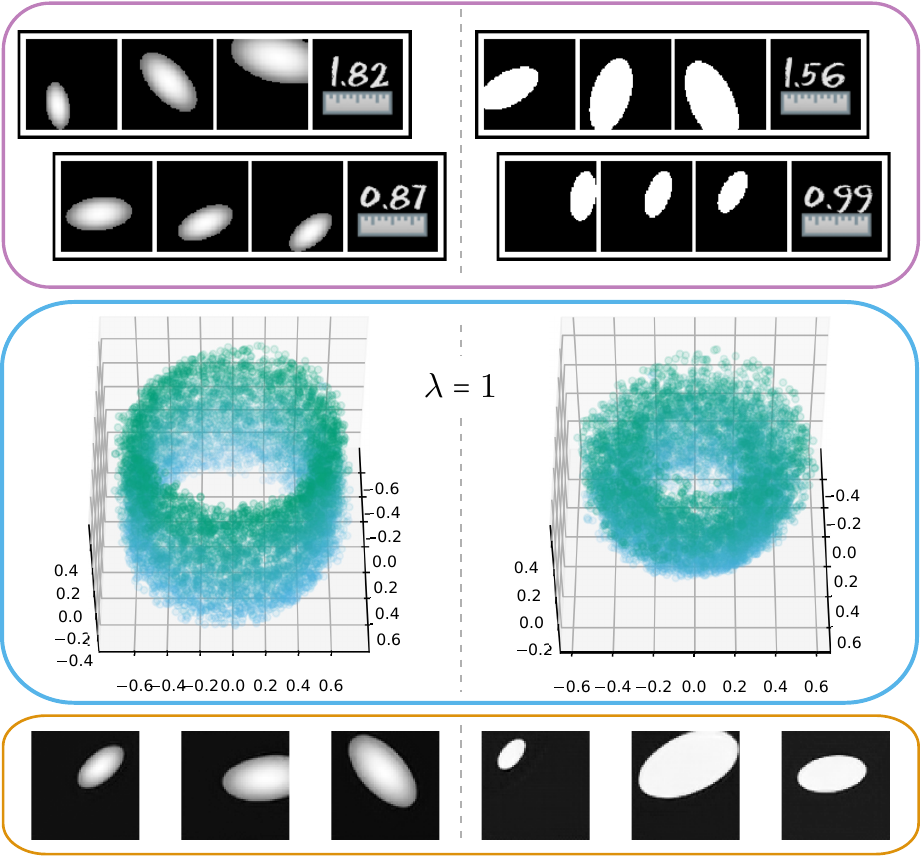}
\caption{Obtained results for anisotropic Gaussian dataset (G); the shown latent manifold dimensions represent rotation ($x$-$y$-axis) and scale ($z$-axis).
   The noise in the right experiment stems from quantizing all input images to binary ones. }
    \label{fig:quantization}
\end{figure}

\Cref{fig:sundials-lambda,fig:cows-qualitative,fig:quantization} show further obtained manifold embeddings in latent space for different loss weights and for the datasets (R) and (G).
In all cases we observe smooth embeddings that neatly reproduce the geometry and topology of the underlying manifold $M$.
For visualizing the manifold embeddings $\phi(M)$ we simply perform a principal component analysis (PCA) in latent space and display the resulting top three dimensions
(in \cref{fig:cows-qualitative} a second set of principal components is shown in addition).
To illustrate that our approach allows separate training of encoder and decoder,
for \cref{fig:masterpiece} we first trained the encoder map $\phi$ on its own by minimizing $E^{\sample}(\phi)$
and subsequently trained the decoder map $\psi$ by minimizing, for fixed $\phi$, the \emph{reconstruction loss}
\begin{equation}
R(\phi,\psi)=\frac1{|\sample|}\sum_{(x,y)\in\sample}\|\psi(\phi(x))-x\|_{L^2}^2+\|\psi(\phi(y))-y\|_{L^2}^2
\end{equation}
with $\|\cdot\|_{L^2}$ the $L^2$-norm on images.
For the other datasets we train en- and decoder simultaneously by jointly minimizing $E^{\sample}(\phi)+\kappa R(\phi,\psi)$ for $\phi$ and $\psi$
(where the weight $\kappa>0$ is not expected to have much influence since $\psi$ will try to minimize $R(\phi,\psi)$ anyhow).
We use $\eps=\frac\pi2$ for dataset (S), $\eps=\frac\pi4$ for (R). For (G), for simplicity we used a slightly different sample
$\hat\sample \subset \{(y,y') \in M\times M \,\vert\, d_{S^1}(\alpha, \alpha') \leq \eps\}$ with $\eps=\frac\pi2$.
During training the isometry, flatness, and reconstruction parts of the loss function are all observed to decrease continuously and monotonically (up to the usual stochastic variations)
as shown in \cref{fig:training} for dataset (R), where the full autoencoder is trained simultaneously.
Resulting reconstructions $\psi(\phi(x))$ for random $x\in M$ are exemplarily shown in \cref{fig:masterpiece,fig:quantization} and are of good quality.
\Cref{fig:sundials-lambda} illustrates the effect of the flatness term:
for zero bending penalization $\lambda$ one obtains a standard isometric embedding of $S^2$, while it gets flattened for higher values of $\lambda$
(\cref{fig:masterpiece} shows an intermediate $\lambda$).
Note that extrinsic bending is obviously reduced this way, while nonlinear inplane distortion is increased
(the normal component of the Hessian apparently outweighs the inplane component).

\paragraph{Linear interpolation in latent space.}
\begin{figure}\centering
\includegraphics[]{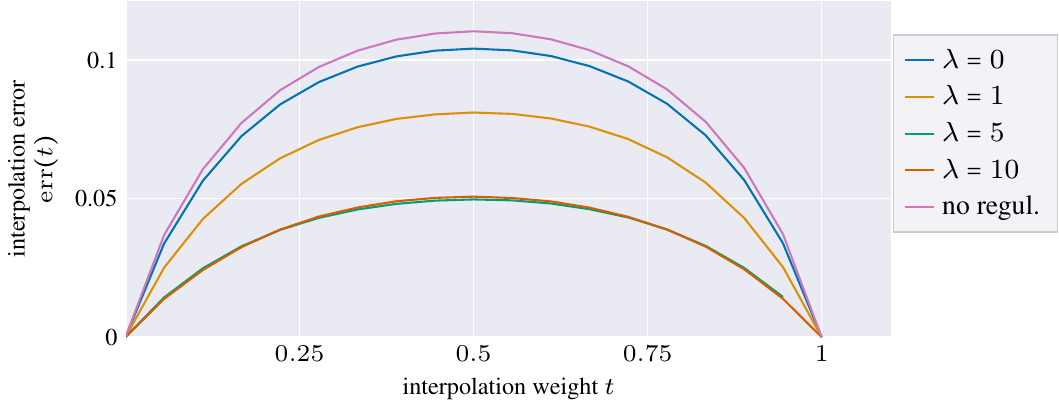}
\caption{\label{fig:cows-eval}Average error of linear interpolation in latent space for dataset (R) and different flatness weights $\lambda$.}
\end{figure}

\Cref{fig:cows-qualitative} illustrates for dataset (R) that the bending term of our loss functional strongly improves the usefulness of linear interpolation in latent space across moderate distances.
While for $\lambda=0$ the decoder output of linear interpolations in latent space does not at all reproduce continuously rotating objects,
it clearly does for bending weight $\lambda=10$.
Let us emphasize that the decoder was in no way regularized in these experiments,
in particular it was not trained on any points obtained via linear interpolation of codes in latent space!
Still the rotated cow is cleanly visible (in contrast to the case $\lambda=0$), having undergone merely some minor smoothing.
Training the decoder additionally on linear interpolants in latent space will naturally improve the results.
We purposely abstained from this extra regularization
since it would allow the decoder to compensate the deficiencies visible for $\lambda=0$
so that the regularizing properties of our encoder loss functional $E^{\sample}$ would be obscured.
For $\lambda=10$ the decoder has to compensate much less and is therefore expected to be more robustly trainable.
We also quantitatively evaluated the quality of linear interpolation in latent space
by measuring the $L^2$-error to the ground truth, geodesic interpolation. We calculate this on a test sample set $\sample'$ as
\begin{align}
\err(t)^2 &= \frac{1}{\abs{\sample'}}\sum_{(x,y)\in\sample'} \err_{\text{i}}(x,y; t)^2 - \err_{\text{b}}(x,y; t)^2\text{ for}\\
\err_{\text{i}}(x,y; t) &= \|\av_M(x,y;t) - \psi(\av_{\R^l}(\phi(x),\phi(y);t))\|_{L^2},\\
\err_{\text{b}}(x,y; t) &= \|\av_M(x,y;t) - \psi(\phi(\av_M(x,y;t)))\|_{L^2},
\end{align}
where $\av_M(x,y;t)$ is the weighted geodesic average of $x,y\in M$ with weights $1-t,t$ and $\av_{\R^l}(a,b;t) = (1-t)a+tb$.
Above, $\err_{\text{i}}$ is the error due to linear interpolation, and $\err_{\text{b}}$ is the base reconstruction error which occurs independently of interpolation.
\Cref{fig:cows-eval} displays $\err(t)$ for different values of $\lambda$, showing a marked error reduction for increasing $\lambda$ up to a saturation around $\lambda=5$.

\paragraph{Additional dimensions exploited by the embedding.} 
In our experiments, we set the latent space dimensionality $l$ to commonly used values.
In particular, we take $l$ substantially larger than would minimally be required for a smooth embedding.
This is reasonable since in applications the intrinsic dimensionality $m$ is generally unknown
and since this allows the encoder to improve on the flatness of the embedding at the expense of using more dimensions.
For example, the flat torus should better be embedded as $S^1\times S^1\subset\R^4$ than as a torus in $\R^3$.
\Cref{fig:cows-qualitative} shows that the encoder makes use of that freedom:
even though $l=5$ would be enough ($SO(3)\cong\R P^3$, which embeds into $\R^5$, but not $\R^4$ \cite{Ho40,Ha38}),
the graph of the standard variation along the principal components of latent space shows that 9 Euclidean dimensions are used for the embedding.
For the experiments with datasets (S) and (G), the embedding used 3 and 5 Euclidean dimensions, respectively.

\paragraph{Noise in the embedding due to image quantization.}
To illustrate the regularization properties of our loss functional we avoided sources of noise in our experiments, as those would require additional tailored regularization.
For the anisotropic Gaussian (G) we now illustrate what effect a simple type of noise can have on the embedding $\phi$ without additional regularization:
we simply round all Gaussian images to binary images.
This quantization makes ellipses in nearby positions, orientations and scalings harder to distinguish.
\Cref{fig:quantization} right shows that a cylindrical structure of the resulting latent manifold $\phi(M)$ is still observable,
though it is thickened and much less clean than on the left.

\paragraph{Acknowledgement}
This work was supported by the Deutsche Forschungsgemeinschaft (DFG, German Research Foundation) 
via project 211504053 - Collaborative Research Center 1060 and via Germany's Excellence Strategy project 390685813 - 
Hausdorff Center for Mathematics and  project 390685587 - Mathematics M\"unster: Dynamics-Geometry-Structure.
\FloatBarrier
\bibliographystyle{alpha}
\bibliography{bibliography}
\end{document}